\documentclass[preprint]{imsart}

\RequirePackage[OT1]{fontenc}
\RequirePackage{amsthm,amsmath}
\RequirePackage[numbers]{natbib}
\RequirePackage[colorlinks,citecolor=blue,urlcolor=blue]{hyperref}

\usepackage{float}
\usepackage{amsmath,amssymb,epsfig}
\usepackage{graphics}
\usepackage{tabularx}
\usepackage{framed}
\usepackage[mathscr]{euscript}
\usepackage{imsart}

\usepackage{epstopdf}

\startlocaldefs
\numberwithin{equation}{section}
\theoremstyle{plain}

\long\def\comment#1{}

\newtheorem{theorem}{Theorem}

\theoremstyle{definition}

\newcommand{\eps}{\varepsilon}

\newcommand{\be}{\begin{eqnarray}}
\newcommand{\ee}{\end{eqnarray}}

\newcommand{\thetahat}{\hat \theta}

\newcommand{\ba}{\begin{array}}
\newcommand{\ea}{\end{array}}
\newcommand{\bs}{\begin{align}\begin{split}\nonumber}
\newcommand{\bsnumber}{\begin{align}\begin{split}}
\newcommand{\es}{\end{split}\end{align}}

\renewcommand{\(}{\left(}
\renewcommand{\)}{\right)}
\renewcommand{\[}{\left[}
\renewcommand{\]}{\right]}
\renewcommand{\hat}{\widehat}

\newcommand{\En}{{\mathbb E_n}}

\def\x{{x}}
\def\supp{{\rm support}}

\renewcommand{\hat}{\widehat}
\renewcommand{\leq}{\leqslant}
\renewcommand{\geq}{\geqslant}
\renewcommand{\supp}{\text{supp}}

\endlocaldefs

\begin{document}

\begin{frontmatter}
\title{Sharp Convergence Rates for Forward Regression in High-Dimensional Sparse Linear Models}
\runtitle{Forward Regression}

\begin{aug}
\author{\fnms{Damian} \snm{Kozbur}\ead[label=e1]{damian.kozbur@econ.uzh.ch.}}

\thankstext{t1}{ This version is of  \today.   An earlier version of this paper, \textit{Testing-Based Forward Model Selection} \cite{TBFMS1},  is being split into two papers.  The current paper presents fundamental results needed for analysis of forward regression in general settings, while the other paper focuses on using hypothesis tests rather than a simple threshold to decided which covariates enter the selected model.   I gratefully acknowledge helpful discussion with Christian Hansen, Tim Conley, Attendants at the ETH Z\"urich Seminar f\"ur Statistik Research Seminar, Attendants at the Center for Law and Economics Internal Seminar, as well as financial support of the ETH Fellowship program.}

\runauthor{Damian Kozbur}

\affiliation{University of Z\"urich }

\address{University of Z\"urich \\ Department of Economics \\ Sch\"onberggasse 1, 8006 Z\"urich\\
\printead{e1}\\
}

\end{aug}

\begin{abstract}
 Forward regression is a statistical model selection and estimation procedure which inductively selects covariates that add predictive power into a working statistical regression model.   Once a model is selected, unknown regression parameters are estimated by least squares.  This paper analyzes forward regression in high-dimensional sparse linear models.  Probabilistic bounds for prediction error norm and number of selected covariates are proved.  The analysis in this paper gives sharp rates and does not require $\beta$-min or irrepresentability conditions.   \end{abstract}


\begin{keyword}
\kwd{ forward regression, high-dimensional models, sparsity,  model selection \\ \ \\ \textbf{Import bibliographic note.} \ The content presented in this paper has been merged into another paper titled \textit{Testing-Based Forward Model Selection}, with a draft available at https://arxiv.org/abs/1512.02666.  The two pervious papers were originally separate projects, but have now been merged in preparation for the publication process.  No original material is submitted simultaneously to multiple peer-reviewed journals.  \\ \ \\ This paper remains posted on ArXiv for the time being in order to serve as a record of the progression of drafts posted to the internet}
\end{keyword}

\end{frontmatter}

\section{Introduction}

Forward regression is a statistical model selection and estimation technique that inductively selects covariates which substantially increase predictive accuracy into a working statistical model until a stopping criterion is met.  Once a model is selected, unknown regression parameters are estimated by least squares. 
This paper studies statistical properties and proves convergence rates for forward regression in high-dimensional settings.  

Dealing with a high-dimensional dataset necessarily involves dimension reduction or regularization.  A principal goal of research in high-dimensional statistics and econometrics is to generate predictive power that guards against false discovery and overfitting, does not erroneously equate in-sample fit to out-of-sample predictive ability, and accurately accounts for using the same data to examine many different hypotheses or models.  Without dimension reduction or regularization, however, any statistical model will overfit a high dimensional dataset.  Forward regression is a method for doing such regularization which is simple to implement, computationally efficient, and easy to understand mechanically.  

There are several earlier analyses of forward selection.  \cite{Wang:UltraHighForwardReg}  gives bounds on the performance and number of selected covariates under a $\beta$-min condition which restricts the minimum magnitude of nonzero regression coefficients.  \cite{Zhang:GreedyLeastSquares} and \cite{Tropp:Greed} prove performance bounds for greedy algorithms under a strong irrepresentability condition, which restricts the empirical covariance matrix of the predictors.   \cite{Submodular:Spectral} prove bounds on the relative performance in population $R$-squared of forward regression (relative to infeasible $R$-squared) when the number of variables allowed for selection is fixed.   

A key difference between the analysis in this paper relative to previous analysis of forward regression is that all bounds are stated in terms of the sparse eigenvalues of the empirical Gram matrix of the covariates.  No $\beta$-min or irrepresentability conditions are required.  Under these general conditions, this paper proves probabilistic bounds on the predictive performance which rely on a bound on the number of selected covariates.  In addition, the rates derived here are sharp.  

A principal idea in the proof is to track average correlation among selected covariates.   The only way for many covariates to be falsely selected into the model is that they be correlated to the outcome variable.  Then, by merit of being correlated to the outcome, subsets of the selected covariates must also exhibit correlation amongst each other.  On the other hand, sparse eigenvalue conditions on the empirical Gram matrix put upper limits on average correlations between covariates.  These two observations together imply a bound on the number of covariates which can be selected.   Finally, the convergence rates for forward regression follow.


A related method is forward-backward regression, which proceeds similarly to forward regression, but allows previously selected covariates to be discarded from the working model at certain steps. The convergence rates proven in this paper match those in the analysis of a forward-backward regression in \cite{TongZhang:ForwardBackward}.  Despite the similarity between the two procedures, it is still desirable to have a good understanding of forward selection.  An advantage of forward selection relative to forward-backward is computational simplicity.  In addition, understanding the properties of forward selection may lead to better understanding of general early stopping procedure in statistics (see \cite{Yao2007:EarlyStopping:GradDesc}, \cite{zhang2005} ) as well as other greedy algorithms (see \cite{BuhlmannBoosting2006},  \cite{Freund+Schapire:1996}  ).  The analysis required for forward regression requires quite different techniques, since there is no chance to correct `model selection mistakes.'   

There still are many other sensible approaches to high dimensional estimation and regularization.  An important and common approach to generic high dimensional estimation problems is the Lasso.  The Lasso minimizes a least squares criteria augmented with a penalty proportional to the $\ell_1$ norm of the coefficient vector.  For theoretical and simulation results about the performance of Lasso, see \cite{FF:1993} \cite{T1996}, \cite{elements:book}, \cite{CandesTao2007}, \cite{BaiNg2008}, \cite{BaiNg2009b}, \cite{BickelRitovTsybakov2009},  \cite{BuneaTsybakovWegkamp2006}, \cite{BuneaTsybakovWegkamp2007b} \cite{BuneaTsybakovWegkamp2007}, \cite{CandesTao2007}, \cite{horowitz:lasso}, \cite{knight:shrinkage}, \cite{Koltchinskii2009}, \cite{Lounici2008}, \cite{LouniciPontilTsybakovvandeGeer2009}, \cite{MY2007}, \cite{RosenbaumTsybakov2008}, \cite{T1996}, \cite{vdGeer}, \cite{Wainright2006}, \cite{ZhangHuang2006}, \cite{BC-PostLASSO},  \cite{BuhlmannGeer2011}, \cite{BC-PostLASSO}, among many more. 
In addition,\cite{EfronHastieJohnstoneTibshirani2004} have shown that under restrictive conditions, Lasso and forward regression yield approximately the same solutions (see also \cite{MRY2007}).  
This paper derives statistical performance bounds for forward selection which match those given by Lasso in more general circumstances.

Finally, an important potential application for forward regression is as an input for post-model-selection analysis. 
One example is the selection of a conditioning set, to properly control for omitted variables bias when there are many potential control variables (see \cite{BCH-PLM}, \cite{vdGBRD:AsymptoticConfidenceSets}, \cite{BCHK:Panel}).   Another example is the selection of instrumental variables for later use in a first stage regression (see \cite{BellChenChernHans:nonGauss}).   Both applications require a model selection procedure with the hybrid property of both producing a good fit and returning a sparse set of covariates.  The results derived in this paper are relevant for both objectives, deriving bounds for both prediction error norm as well as the size of the selected set for forward regression.

\section{Framework}

The observed data is given by $\mathscr D = \{(x_i,y_i)\}_{i=1}^n$. The data consists of a set of covariates $x_i\in \mathbb R^p,$ as well as outcome variables $y_i \in \mathbb R$ for each observation $i=1,...,n$.  The data satisfy  
$$y_i = x_i'\theta_0 + \eps_i$$
for some unknown parameter of interest $\theta_0 \in \mathbb R^p$ and unobserved disturbance terms $\eps_i \in \mathbb R$.  The covariates $x_i$ are normalized so that $\En[x_{ij}]=0$ and $ \En[x_{ij}^2]=1$ for every $j=1,...,p$, where $\En[ \hspace{.6mm} \cdot \hspace{.6mm} ] = \frac{1}{n}\sum_{i=1}^n ( \hspace{.2mm} \cdot \hspace{.2mm}  )$ denotes empirical expectation.  Finally, the parameter $\theta_0$ is sparse in the sense that the set of non-zero components of $\theta_0$, denoted $S_0=\text{supp}(\theta_0)$, has cardinality $s_0<n$. The interest in this paper is to study how well forward regression can estimate $x_i'\theta_0$ for $i=1,...,n$.

Define a loss function $\ell(\theta)$ $$\ell(\theta) = \En[ (y_i - x_i'\theta)^2].$$
 Note that $\ell(\theta)$ depends on $\mathscr D$, but this dependence is suppressed from the notation.  
Define also $$\ell(S) = \min_{\theta: \text{supp}(\theta) \subseteq S} \ell(\theta).$$The estimation strategy proceeds by first searching for a sparse subset $\hat S \subseteq \{1,...,p\}$, with cardinality $\hat s$,  that assumes a small value of
$\ell(S)$,
followed by estimating $\theta_0$ with least squares via

$$\hat \theta \in \text{arg} \min_{\theta:\text{supp}(\theta) \subseteq \hat S} \ell(\theta).$$
This gives the construction of the estimates $ x_i'\hat \theta$ for $i=1,...,n$.
The paper provides bounds for the prediction error norm defined by $$\En [(x_i' \theta_0 - x_i'\hat \theta )^2]^{1/2}. $$

The set $\hat S$ is selected by forward regression.    For any $S$ define the incremental loss from the $j$th covariate by

$$\Delta_j \ell(S) = \ell(S \cup \{j\} ) -   \ell(S).$$ 

\noindent Consider the greedy algorithm which inductively selects the $j$th covariate to enter a working model if $-\Delta_j\ell(S)$ exceeds a threshold $t$:  $$-\Delta_j\ell(S) > t$$ and $\Delta_j\ell(S) \geq \Delta_k\ell(S)$ for each $k \neq j$.    The threshold $t$ is chosen by the user; it is the only tuning parameter required.  This defines forward regression.  It is summarized formally here:

\

\begin{framed}

\

\centerline{{Algorithm 1: Forward Regression}}

\

\noindent \textbf{Initialize.}  \noindent Set $\hat S = \{\}$.

\noindent \textbf{For $1 \leq k \leq p$:} 

\textbf{If:} $-\Delta_j\ell(S) > t$ for some $j\in \{1,...,p\} \setminus \hat S$, then select $$\hat j \in \text{arg} \max  \left \{-\Delta_j\ell(S) : -\Delta_j\ell(S) > t \right \}.$$ \hspace{ 2.8mm} \textbf{Update: }  $\hat S = \hat S \cup \{\hat j\}$. 

\textbf{Else:}  \textbf{Break.}

\noindent \textbf{Set: } $$\hat \theta \in \arg \min_{\theta:\supp(\theta) \subset \hat S} \ell({\theta}).$$

\end{framed}

\section{Analysis of Forward Regression}

In order to state the main theorem, a few more definitions are convenient.  Define the empirical Gram matrix $G_x$ by $G_x = \En[x_ix_i']$. Let $\varphi_{\min}(s)(G_x)$ denote the minimum $s$-sparse eigenvalues given by $$\varphi_{\min}(s)(G_x) = \min_{S \subseteq \{1,...,p\}: |S| \leq s} \lambda_{\min}([G_x]_{S,S})$$ where $[G_x]_{S,S}$ is the principal submatrix of $G_x$ corresponding to the component set $S$.   
Let 
$$  \mathscr C_1 = \sqrt{\hat s + s_0}  \varphi_{\min}(\hat s + s_0)(G_x)^{-1} \[2\| \En[\eps_i x_i'] \|_\infty +t^{1/2} \].$$
For each positive integer $m$, let 
\begin{align*}
\mathscr C_2(m) & = 1+ 72  \times 1.783^2  \times  \varphi_{\min}(m+s_0)(G_x)^{-5}.
\end{align*}

The  above quantities are useful for displaying results in Theorem 1. Slightly tighter but messier usable quantities than $\mathscr C_1$ and $\mathscr C_2(m)$ are derived in the proof. Note also that $\mathscr C_1$ depends on $\hat s$.

\

\begin{theorem}

Consider data $\mathscr D$ with parameter $\theta_0$.  Then under Algorithm 1 with threshold $t$,

$$
\En [(x_i'\theta_0 -\x_{ i}'\hat \theta)^2]^{1/2} \leq \mathscr C_1. $$
For every integer $m \geq 0$ such that  ${t^{1/2}} \geq  2\varphi_{min}(m+s_0)(G_x)^{-1} \| \En[x_i \eps_i]\|_{\infty}$ and $m \leq |\hat S \setminus S_0|$, it holds that 
$$ \ m \leq \mathscr C_2(m) s_0.$$

\end{theorem}

\

The above theorem calculates explicit constants bounding the prediction error norm.  It is also helpful to consider the convergence rates implied by Theorem 1 under more concrete conditions on $\mathscr D$.  Next, consider the following conditions on a sequence of datasets $\mathscr D_n$.  In what follows, the parameters $\theta_0$, the thresholds $t$, and distribution of the data can all depend on $n$.

\

\noindent \textbf{Condition 1} [\textit{Model and Sparsity}]. $s_0 = o(n)$.

\

\noindent \textbf{Condition 2} [\textit{Sparse Eigenvalues}]. There is a sequence $K_n$ such that $s_0= o(K_n) $.   In addition, $\varphi_{\min}(K_n)(G_x)^{-1} = O(1)$ with probability $1-o(1)$. 

\

\noindent \textbf{Condition 3} [\textit{Threshold and Disturbance Terms}]. The threshold satisfies $t = O({ \log p /n})$.  In addition, ${t^{1/2}} \geq 2 \varphi_{\min}(K_n)(G_x)^{-1} \| \En[x_i \eps_i]\|_{\infty}$
 with probability $1-o(1)$.

\begin{theorem}
For a sequence of datasets $\mathscr D_n$ with parameters $\theta_0$ and thresholds $t$ satisfying Conditions 1-3,  the bounds
$$
\En [(x_i'\theta_0 -\x_{ i}'\hat \theta)^2]^{1/2} = O(\sqrt{ s_0 \log p /n }),$$
$$ \ \hat s \leq O(1)s_0$$
hold with probability $1-o(1)$. 
\end{theorem}

\

The theorem shows that forward regression exhibits asymptotically the same convergence rates in prediction error norm as other high-dimensional estimators like Lasso, provided an appropriate threshold $t$ is used.  In addition, forward regression selects a set with cardinality commensurate with $s_0$.  

Condition 1 bounds the size of $S_0$ and requires that the sparsity level is small relative to the sample size.
Condition 2 is a sparse eigenvalue condition useful for proving results about high dimensional techniques like Lasso.  In standard regression analysis where the number of covariates is small relative to the sample size, a conventional assumption used in establishing desirable properties of conventional estimators of $\theta$ is that $G_x$ has full rank. In the high-dimensional setting,  $G_x$ will be singular if $p>n$ and may have an ill-behaved inverse even when $p \leq n$. However, good performance of many high-dimensional estimators only requires good behavior of certain moduli of continuity of $G_x$.   There are multiple formalizations and moduli of continuity that can be considered here; see  \cite{BickelRitovTsybakov2009}. This analysis focuses on a simple eigenvalue condition which was used in \cite{BellChenChernHans:nonGauss}.  Condition 2 could be shown to hold under more primitive conditions by adapting arguments found in \cite{BC-PostLASSO} which build upon results in \cite{ZhangHuang2006} and \cite{RudelsonVershynin2008}; see also \cite{RudelsonZhou2011}.  Condition 2 is notably weaker than previously used irrepresentability conditions.  Irrepresentability conditions require that for certain sets $S$ and $k \notin S$, letting $x_{iS}$ be the subvector of $x_i$ with components $j\in S$, that $\| \En[x_{iS}x_{iS}']^{-1}\En[x_{iS}x_{ik}'] \|_1$ is bounded, or even strictly less than 1.

Condition 3 is a regularization condition similar to regularization conditions common in the analysis of Lasso.  The condition, requires $ t^{1/2}$ to dominate a multiple of the $\| \En[x_i \eps_i]\|_\infty$.  This condition is stronger than that typically encountered with Lasso, because the multiple relies on the sparse eigenvalues of $G_x$.  To illustrate why such a condition is useful, let $\check x_{ij}$ denote $x_{ij}$ residualized away from previously selected regressors and renormalized.  Then even if  $\En[x_{ij} \eps_i] < t^{1/2}$,  $\En[\check x_{ij}\eps_i]$ can exceed $t^{1/2}$ resulting in more selections into the model.   Nevertheless, using the multiple $2 \varphi_{\min}(K_n)(G_x)^{-1}$ which stays bounded with $n$, is sufficient to ensure that $\hat s$ does not grow faster than $s_0$.  From a practical standpoint, this condition also requires the user to know more about the design of the data in choosing an appropriate $t$.  Choosing feasible thresholds which satisfy a similar condition to Condition 3 is considered in \cite{TBFMS1}.

\

\begin{theorem} For a sequence of datasets $\mathscr D_n$ with parameters $\theta_0$ and thresholds $t$ satisfying Conditions 1-3,  the bounds  $$\| \theta_0 - \hat \theta \|_2 = O(\sqrt{ s_0 \log p /n} ) \text{ and }\| \theta_0 - \hat \theta \|_1 = O(\sqrt{ s_0^2 \log p /n  }  )$$ hold with probability $1-o(1).$
\end{theorem}  

\

\
Finally, two direct consequence of Theorem 2 are   bounds on the deviations $\| \hat \theta - \theta_0\|_1$ and $  \| \hat \theta - \theta_0\|_2$ of $\hat \theta$ from underlying unknown parameter $\theta_0$.  Theorem 3 above shows that deviations of $\hat \theta$ from $\theta_0$ also achieve rates typically encountered in high-dimensional estimators like Lasso.

\section{Proof of Theorem 1}

\begin{proof}

The proof of Theorem 1 is divided into seven steps.  Step 1 shows the first statement of Theorem 1.  Step 2 defines a useful normalization of the selected covariates.  Step 3 establishes certain bounds on the average correlation between selected covariates.  Steps 4-6 show that if $\hat s$ is too high, then there must exist subsets of the selected covariates over which the average correlation must exceed what is permitted by assumption on the sparse eigenvalues of the empirical Gram matrix $G$.  Step 7 concludes by pulling together the previous six steps.

\subsection*{Step 1} This first section of the proof provides a bound on $ \En [ (x_i'\theta_0-x_i'\hat \theta )^2] $ which depends on $\hat s$ thereby proving the first statement of Theorem 1.  First note that
$\ell(\hat S) = \ell(\hat S \cup S_0) + [\ell(\hat S) - \ell(\hat S \cup S_0)]$.
Note that $\ell(\hat S) = \ell(\thetahat)$ and $\ell(\hat S \cup S_0) \leq \ell(\theta_0)$. In addition, by
Lemma 3.3 of \cite{Submodular:Spectral},  

$$\ell(\hat S) - \ell(\hat S \cup S_0) \leq \varphi_{\min}(\hat s + s_0)(G)^{-1} \sum_{j \in  S_0 \setminus \hat S} (-\Delta_j \ell(\hat S)) \leq s_0t\varphi_{\min}(\hat s+ s_0)(G)^{-1}.$$  

\noindent This gives
$$\ell(\hat \theta) \leq \ell(\theta_0) + s_0t\varphi_{\min}(\hat s + s_0)(G)^{-1}.$$ 

\noindent Expanding the above two quadratics in $\ell(\cdot)$ gives
\begin{align*} \En [ (x_i'\theta_0-x_i'\hat \theta )^2] &\leq |2\En[\eps_i x_i'(\hat \theta - \theta_0)] |+ s_0t\varphi_{\min}(\hat s + s_0)(G)^{-1}\\
& \leq 2\| \En[\eps_i x_i'] \|_\infty \|\theta_0 - \hat \theta \|_1 + s_0t\varphi_{\min}(\hat s+s_0)(G)^{-1}
\end{align*}
To bound $\| \theta_0 - \hat \theta \|_1$:
\begin{eqnarray*}
&\| \theta_0 - \thetahat \|_1 &\leq \sqrt{\hat s + s_0} \| \theta_0 - \thetahat \|_2\\ 
&&\leq \sqrt{\hat s + s_0} \varphi_{\min}(\hat s + s_0)(G)^{-1} \En[(x_i'\theta_0 - x_i' \thetahat)^2]^{1/2} .  \\
\end{eqnarray*}
Combining the above bounds and dividing by $\En[(x_i'\theta_0 - x_i' \thetahat)^2]^{1/2} $  gives 
\begin{align*}\En [ (x_i'\theta-x_i'\hat \theta )^2]^{1/2} &\leq    2\| \En[\eps_i x_i'] \|_\infty \sqrt{\hat s + s_0} \varphi_{\min}(\hat s + s_0)(G)^{-1} \\ & + \frac{s_0t\varphi_{\min}(\hat s + s_0)(G)^{-1}}{\En[(x_i'\theta_0 - x_i' \thetahat)^2]^{1/2}  }.
\end{align*}
Finally, either $\En[(x_i'\theta_0 - x_i' \thetahat)^2]^{1/2} \leq \sqrt{  s_0t\varphi_{\min}(\hat s + s_0)(G)^{-1} }$, in which case the first statement of Theorem 1 holds,  or alternatively $\En[(x_i'\theta_0 - x_i' \thetahat)^2]^{1/2} > \sqrt{  s_0t\varphi_{\min}(\hat s + s_0)(G)^{-1} }$, in which case 
\begin{align*}\En [ (x_i'\theta-x_i'\hat \theta )^2]^{1/2} &\leq    2\| \En[\eps_i x_i'] \|_\infty \sqrt{\hat s + s_0} \varphi_{\min}(\hat s + s_0)(G)^{-1} \\ & + \sqrt{s_0t\varphi_{\min}(\hat s + s_0)(G)^{-1}}
\end{align*} 
and the first statement of Theorem 1 follows.

\subsection*{Step 2}
This section of the proof defines \textit{true} and \textit{false} covariates, introduces a convenient orthogonalization of all selected covariates, and associates to each false selected covariate a parameter $\tilde \gamma_j$ on which the analysis is based.  

Let $x_j = [x_{1j},...,x_{nj}]'$ be the vector in $\mathbb R^n$ with components $x_{ij}$ stacked vertically.  Similarly, define $\eps = [\eps_1,...,\eps_n]'$ and $y = [y_1,...,y_n]'$.  Let $v_k \in \mathbb R^n$, $k=1,...,s_0$ denote \textit{true covariates} which are defined as the the vectors $x_j$  for $j \in S_0$.  Define \textit{false covariates} simply as those which do not belong to $S_0$.  

Consider any point in time in the the forward regression algorithm when there are $m$ false covariates selected into the model.  These falsely selected covariates are denoted $w_1,...,w_m$, each in $\mathbb R^n$, ordered according to the order they were selected.

  The true covariates are also ordered according to the order they are selected into the model.  Any true covariates unselected after the $m$ false covariate selection are temporarily  ordered arbitrarily at the end of the list.   Let $\mathscr M_{k}$ be projection in $\mathbb R^n$ onto the space orthogonal to $\text{span}(\{v_1,...,v_k\})$.  Let $$\tilde v_k = \frac{ \mathscr M_{k-1}v_k }{ (v_k'\mathscr M_{k-1}v_k)^{1/2}} \ \text{for} \ k = 1,...,s_0.$$  
In addition,  set $$\tilde \eps =  \frac{\mathscr M_{s_0 }\eps}{(\eps'\mathscr M_{s_0}\eps)^{1/2}}.$$

Let $\tilde V_{\text{temp}} = [\tilde v_1,...,\tilde v_{s_0}]$, ordered according to the temporary order.  Note that there is $\tilde \theta \in \mathbb R^{s_0}$ and $\tilde \theta_{\tilde \eps} \in \mathbb R$ such that $$\tilde V_{\text{temp}}   \tilde \theta_{\text{temp}} + \tilde \theta_{\tilde \eps} \tilde \eps = y.$$

At this time, reorder the true covariates.  Let $\hat k$ denote the index of the final true covariate selected into the model when the $m$-th false covariate is selected.  The variables $\tilde v_1,...,\tilde v_{\hat k}$ maintain their original order.  The unselected true covariates $\tilde v_{\hat k+1},..., \tilde v_{s_0}$ are reordered in such a way that under the new ordering, $\tilde \theta_{k, \text{temp}} \geq \tilde \theta_{l, \text{temp}}$ whenever $l > k$.   Also define $ \tilde V = [\tilde v_1,...,\tilde v_{s_0}]$ consistent with the new ordering.  Redefine $\tilde \theta$ by $\tilde V \tilde \theta + \tilde \theta_{\tilde \eps} \tilde \eps = y$ so that it is also consistent with the new ordering.  Note that no new orthogonalization needs to be done.  


For any set $S$, Let $\mathscr Q_{S}$ be projection onto the space orthogonal to $\text{span}(   \{   x_j, \ j \in S \}  )$.  For each selected covariate, $w_j$, set  $S_{\text{pre-}w_j}$ to be the set of (both true and false) covariates selected prior to $w_j$.  Define 
$$\tilde w_j = c_j \mathscr Q_{S_{\text{pre-}w_j}}w_j $$
where the normalization constants $c_j$ are defined in the next paragraph.

 Each $\tilde w_j$ can be decomposed into components $\tilde w_j = \tilde r_j + \tilde u_j$ with $\tilde r_j \in \text{span}( \tilde V)$ and  $\tilde u_j \in \text{span}( \tilde V)^\perp$.  The normalizations $c_j$ introduced above are then chosen so that $\tilde u_j' \tilde u_j  = 1$.

Associates to each false covariate $\tilde w_j$, a vector $\tilde \gamma_j \in \mathbb R^{s_0}$, defined as the solution in $\mathbb R^{s_0}$ to the following equation $$\tilde V  \tilde \gamma_j=\tilde r_j.$$  Set $\tilde \gamma_{j\tilde \eps} = \tilde \eps' \tilde w_j$.   Assume without loss of generality that each component of $\tilde \theta$ is positive (since otherwise, the true covariates can just be multiplied by $-1$.)  Also assume without loss of generality that $\tilde \gamma_{j}'\tilde \theta \geq 0$.  


\subsection*{Step 3}  

This section provides upper bounds on quantities related to the $\tilde \gamma_{j}$ defined above.  The idea guiding the argument in the next sections is that if too many covariates $w_j$ are selected, then on average they must be  correlated with each other since they must be correlated to $y$.  For a discussion of partial transitivity of correlation, see \cite{tao:blog:transitivity}.  If the covariates are highly correlated amongst themselves, then $\varphi_{\min}( m + s_0 )(G)^{-1}$ must be very high.  As a result, the sparse eigenvalues of $G$ can be used to upper bound the number of selections.  Average correlations between covariates are tracked with the aid of the quantities $\tilde \gamma_j$.

Divide the set of false covariates into two sets $A_1$ and $A_2$ where
$$A_1 = \left \{ j : | \tilde \gamma_{j \tilde \eps} | \leq  \frac{t^{1/2}n^{1/2}}{ (2\eps'\mathscr M_{s_0} \eps )^{1/2}}  \right \}, \ A_2 = \left \{ j : | \tilde \gamma_{j \tilde \eps} |>   \frac{t^{1/2}n^{1/2}}{ (2\eps'\mathscr M_{s_0} \eps )^{1/2}}  \right \}.$$ Sections 3 - 5 of the proof bound the number of elements in $A_1$.  Section 6 of the proof bounds the number of elements in $A_2$.  

Suppose the set $A_1$ contains $m_1$ total false selections.  Collect these false selections into
 $\tilde W = [\tilde w_{j_1},...,\tilde w_{j_{m_1}}]$.   Set $\tilde R = [\tilde r_{j_1},...,\tilde r_{j_{m_1}}], \tilde U = [\tilde u_{j_1},...,\tilde u_{j_{m_1}}]$.  Decompose $\tilde W= \tilde R+ \tilde U$. Then $\tilde W'\tilde W =  \tilde  R' \tilde R  +  \tilde U'  \tilde U $.  Since $diag( \tilde U' \tilde U ) = I$, it follows that the average inner product between the $\tilde u_j$, given by $\bar \rho$: $$\bar \rho = \frac{1}{m_1(m_1-1)} \sum_{j \neq l \in A_1} \tilde u_j '\tilde u_l ,$$ must be bounded below by $$\bar \rho \geq -\frac{1}{{m_1}-1}$$ due to the positive definiteness of $ \tilde U' \tilde U $.  This implies an upper bound on the average off-diagonal term in $ \tilde R ' \tilde R $ since $ \tilde W' \tilde W $ is a diagonal matrix.  Since $\tilde v_k$ are orthonormal, the sum of all the elements of $ \tilde R' \tilde R $ is given by $ \| \sum_{j\in A_1} \tilde \gamma_j \|_2^2$.  Since $  \| \sum_{j\in A_1} \tilde \gamma_j   \|^2_2 = \sum_{j\in A_1} \| \tilde \gamma_j' \|^2_2+ \sum_{j \neq l \in A_1} \tilde \gamma_j'\tilde \gamma_l$ and 
since $ \tilde W' \tilde W $ is a diagonal matrix, it must be the case that $$\frac{1}{m_1(m_1-1)}\sum_{j \neq l \in A_1} \tilde \gamma_j'\tilde \gamma_l = -\bar \rho.$$  Therefore,  
$$\bar \rho = \frac{1}{{m_1}(m_1-1)} \( \Big \| \sum_{j\in A_1} \tilde \gamma_j  \Big \|^2_2 - \sum_{j \in A_1} \| \tilde \gamma_j\|^2_2 \)  \leq \frac{1}{m_1-1}.$$
This implies that $$ \Big \| \sum_{j \in A_1}  \tilde \gamma_j  \Big  \|^2_2 \leq {m_1}+ \sum_{j\in A_1} \| \tilde \gamma_j\|^2_2 . $$


Next, bound $\max_{j \in A_1} \| \tilde \gamma_j \|_2^2.$  Note $\| \tilde \gamma_{j} \|_2^2 =  \| \tilde r_j\|_2^2$ since $\tilde V$ is orthonormal.    Note that $\| \tilde u_j \|_2^2 / \| \tilde w_j \|^2_2 = 1/\| \tilde w_j \|_2^2$ is lower bounded by $\varphi_{\min}(m+s_0)(G)$.  This follows from the fact that you can associate $\| \tilde u_j/c_j  \|_2^2$ to an element of a the inverse covariance matrix for $w_j$ and previously selected covariates.  Therefore, $ \| \tilde r_j\|_2^2  = \| \tilde w_j \|_2^2 - 1 \leq \varphi_{\min}(m+s_0)(G)^{-1} - 1$.  It follows that 

$$\max_{j \in A_1} \| \tilde \gamma_j \|_2^2 \leq  \varphi_{\min}(m+s_0)(G)^{-1} -1. $$

\noindent This then implies that $$\Big \| \sum_{j \in A_1}  \tilde \gamma_j  \Big \|^2_2 \leq {m_1} \varphi_{\min}({m}+s_0)(G)^{-1}.$$
The same argument as above also shows that for any choice $e_j \in \{ -1, 1 \}$ of signs,  it is always the case that 
$$\Big \| \sum_{j \in A_1} e_j \tilde \gamma_j  \Big \|^2_2 \leq  {m_1} \varphi_{\min}({m}+s_0)(G)^{-1}.$$
(In more detail, take $\tilde W_e = [\tilde w_{j_1}e_{j_1},...,\tilde w_{j_{m_1}} e_{j_{m_1}}],$ etc. and rerun the same argument.)

\subsection*{Step 4} Next search for a particular choice of signs $\{ e_j \}_{j\in A_1}$ which give a lower bound proportional to ${m_1}^2/s_0$ on the above term.  Note that this will imply an upper bound on $m_1$.   
For each $k=1,...,s_0$, let $A_{1k}$ be the set which contains those $j  \in A_1$ such that $w_j$ is selected before $v_k$, but not before any other true covariate. Note that the sets $A_{1(\hat k+2)},...,A_{1(s_0+1)}$ are set empty if $\hat k <s_0$.  Also, empty sums are set to zero.
Define the following two matrices:


$$\Gamma = \[\begin{array}{cccc} \sum \limits_{j\in A_{11}} \tilde  \gamma_{j1} & \sum \limits_{j \in A_{11}} \tilde \gamma_{j2} & ... & \sum \limits_{j \in A_{11}} \tilde \gamma_{js_0}\\  \\ 0 & \sum \limits_{j\in A_{12}} \tilde \gamma_{j2} & ... & \sum \limits_{j \in A_{12}} \tilde \gamma_{j{s_0}} \\ \\ \vdots & \vdots & \ddots & \vdots \\ \\ 0 & 0 & ... &  \sum \limits_{j \in A_{1{s_0}}} \tilde \gamma_{j{s_0}} \end{array} \],   \ B =  \[\begin{array}{cccc} \frac{\tilde\theta_{1}}{\tilde\theta_{1}} & \frac{\tilde\theta_{2}}{\tilde\theta_{1}} &...& \frac{\tilde\theta_{{s_0}}}{\tilde\theta_{1}} \\  \\\frac{\tilde\theta_{2}}{\tilde\theta_{1}} & \frac{\tilde\theta_{2}}{\tilde\theta_{2}} & ... & \frac{\tilde\theta_{{s_0}}}{\tilde \theta_{2}} \\ \\ \vdots & \vdots & \ddots & \vdots \\ \\ \frac{\tilde\theta_{{s_0}}}{\tilde\theta_{1}} & \frac{\tilde\theta_{{s_0}}}{\tilde\theta_{2}} &...& \frac{\tilde\theta_{{s_0}}}{\tilde\theta_{{s_0}}}   \end{array} \] $$

\

 Note that the $k$th row of $\Gamma$ is equal to $\sum_{j \in A_{1k} }\tilde \gamma_k$ since the orthogonalization process had enforced $\tilde \gamma_{jl} = 0$ for each $l <k$. Therefore, the diagonal elements of the product $\Gamma B$ satisfy the equality 
$$[\Gamma B]_{k,k} = \sum_{j \in A_{k}} \tilde \gamma_j ' \tilde \theta/\tilde \theta_{k}.$$
Let $C_1, C_2$ be constants such that 
$$\tilde \gamma_j ' \tilde \theta / \tilde \theta_k \geq C_1$$  for  $j \in A_{1k}$, and $$\tilde \theta_k / \tilde \theta_l \geq C_2$$  for $l >k.  $ 
These key constants are calculated explicitly in Section 5 of the proof.  They imply that 
$$[\Gamma B]_{k,k} \geq C_1|A_{1k}|  \ \ \text{and} \ \ \text{tr}(\Gamma B) \geq C_1m_1.$$
\noindent Further observe that whenever $\tilde \theta_k \geq C_2 \tilde \theta_l$ for each $k,l>k$, assuming without loss of generality that $C_2 \leq 1$, that $(B + C_2^{-1}I)$  is positive semidefinite.  This can checked by constructing auxiliary random variables who have covariance matrix $B + C_2^{-1}I$:  inductively build a covariance matrix where the $(k+1)$th random variable has $\tilde \theta_k/ \tilde \theta_{k-1}$ covariance with the $k$th random variable.   Then $B + C_2^{-1}I$ has a positive definite symmetric matrix square root so let $D^2 = B + C_2^{-1}I$.  Therefore, $B = (D + C_2^{-1/2}I ) (D - C_2^{-1/2}I). $ Note that the rows (and columns) of $D$ each have norm $\leq 1+C_2^{-1}$ and therefore $B$ decomposes into a product $B = E'F$ where the rows of $E,F$ have norms bounded by $1 + C_2^{-1} + C_2^{-1/2}$.  Therefore, let $C_3 = 1 + C_2^{-1} + C_2^{-1/2}$.  

Consider the set $$\mathscr G_{s_0} = \{ Z \in \mathbb R^{s_0 \times s_0} : Z_{ij} = X_i'Y_j  \text{ for some} \ X_i,Y_j \in \mathbb R^{s_0}, \| X_i \|_2 ,\|Y_j \|_2 \leq1 \}$$
\noindent and observe that $\bar B : = {C_3}^{-1} B \in \mathscr G_{s_0} .$
Then this observation allows the use of Grothendieck's inequality (using the exact form described in \cite{Vershynin:Le:GraphConcentration}) which gives 
$$\max_{Z \in \mathscr G_{s_0}} \text{tr}( \Gamma Z)\leq K_G^{\mathbb R}  \| \Gamma' \|_{\infty \rightarrow 1}.  $$

\noindent Here, $ K_G^{\mathbb R} $ is an absolute constant which is known to be less than 1.783.  It does not depend on $s_0$. Therefore,  $C_1 m  \leq \text{tr}(\Gamma B) = C_3 \text{tr}(\Gamma \bar B)\leq \max_{Z \in \mathscr G_{s_0}}\text{tr}( \Gamma Z)\leq K_G^{\mathbb R}  \| \Gamma' \|_{\infty \rightarrow 1}$, which implies  
$$\({K_G^{\mathbb R}}\) ^{-1} {C_3}^{-1} C_1m_1\leq  \| \Gamma ' \|_{\infty \rightarrow 1}.$$
\noindent Therefore, there is $\nu \in \{-1,1\}^{s_0}$ such that $\|\nu' \Gamma\|_1 \geq \(K_G^{\mathbb R} \)^{-1}{C_3}^{-1} C_1m_1$.  For this particular choice of $\nu$, it follows that $$ \|\nu' \Gamma \|_2 \geq s_0^{-1/2} \(K_G^{\mathbb R} \)^{-1}{C_3}^{-1} C_1 m_1.$$

\noindent Then by definition of $\Gamma$, $\|\nu'\Gamma\|_2^2 = \| \sum_{k=1}^{s_0} \sum_{j \in A_{1k}} \nu_k \tilde \gamma_{j}  \|^2_2$.
In Section 3, it was noted that $ \| \sum_{j=1}^{m_1}  e_j \tilde \gamma_{j}  \|^2_2 \leq  m_1 \varphi_{\min}(m+s_0)(G)^{-1}$ for any choice of signs $e_j \in \{-1,1\}^{m_1}$.  It follows that 
$$s_0^{-1} \(K_G^{\mathbb R} \)^{-2}{C_3}^{-2} C_1^2 m_1^2 \leq m_1\varphi_{\min}(m+s_0)(G)^{-1}$$
\noindent which yields the conclusion $$m_1 \leq  \varphi_{\min}(m+s_0)(G)^{-1} C_1^{-2} {C_3}^2 \(K_G^{\mathbb R} \)^2s_0 .$$

\subsection*{Step 5} It is left to calculate $C_1, C_2$ which lower bound  $\tilde \gamma_j ' \tilde \theta / \tilde \theta_k $ for $j \in A_{1k}$ and $ \tilde \theta_k / \tilde \theta_l $ for $l >k$. 
A simple derivation can be made to show that the incremental decrease in empirical loss from the $j$th false selection is

$$-\Delta_j \ell(S_{\text{pre-}w_j}) =  \frac{1}{n} y' \tilde w_j (\tilde w_j' \tilde w_j)^{-1} \tilde w_j'y = \frac{1}{n} \frac{1}{\tilde w_j ' \tilde w_j} (\tilde \theta ' \tilde \gamma_j + \tilde \theta_{\tilde \eps}'\tilde \gamma_{j\tilde \eps})^2$$
Note the slight abuse of notation in $-\Delta_j(S_{\text{pre-}w_j})$ signifying change in loss under inclusion of $w_j$ rather than $x_j$. Next, 
$$ (\tilde \theta ' \tilde \gamma_j + \tilde \theta_{\tilde \eps}'\tilde \gamma_{j\tilde \eps})^2 \leq 2(\tilde \theta ' \tilde \gamma_j )^2 + 2 ( \tilde \theta_{\tilde \eps}'\tilde \gamma_{j\tilde \eps})^2$$
Since $\tilde \theta_{\tilde \eps} =( \eps'\mathscr M_{s_0} \eps)^{1/2}$, $\tilde w_j'\tilde w_j \geq 1$, and $j \in A_1$ it follows that $$\frac{1}{n}\frac{1}{\tilde w_j ' \tilde w_j} ( \tilde \theta_{\tilde \eps}'\tilde \gamma_{j\tilde \eps})^2 \leq \frac{1}{n}  \frac{1}{\tilde w_j ' \tilde w_j}  \tilde \theta_{\tilde \eps}^2\( \frac{t^{1/2}n^{1/2}}{2 (\eps'\mathscr M_{s_0} \eps )^{1/2}}  \)^2 \leq \frac{t}{4}.$$
This implies
$$ \frac{1}{2}( -\Delta_j \ell(S_{\text{pre-}w_j})  )\leq  \frac{1}{n} \frac{1}{\tilde w_j ' \tilde w_j} (\tilde \theta'  \tilde \gamma_j)^2 + \frac{t}{4}.$$
By the condition that the false $j$ is selected, it holds that $ -\Delta_j \ell(S_{\text{pre-}w_j})  > t$ and so $ \frac{1}{4}(-\Delta_j \ell(S_{\text{pre-}w_j}) )> \frac{t}{4}$ which implies that 

$$ \frac{1}{2}(-\Delta_j \ell(S_{\text{pre-}w_j})) - \frac{t}{4}  \geq \frac{1}{4}(-\Delta_j \ell(S_{\text{pre-}w_j})).$$
Finally, this yields that
$$\frac{1}{n\tilde w_j'\tilde w_j} (\tilde \gamma_j ' \tilde \theta)^2  \geq  \frac{1}{4}(- \Delta_j \ell(S_{\text{pre-}w_j})).$$
By the fact that $w_j$ was selected ahead of $v_k$ it holds that 
$$- \Delta_j \ell(S_{\text{pre-}w_j}) \geq - \Delta_k \ell(S_{\text{pre-}w_j}).$$  
Therefore, further bound the righthand side.  Let $\tilde z_k$ be the projection of $\tilde v_k$ onto the space orthogonal to all previously selected (true and false) covariates.  Then 
$$ - \Delta_k \ell(S_{\text{pre-}w_j}) \geq \frac{1}{n} \tilde z_k'\tilde z_k \tilde \theta_k^2.$$
Furthermore, $\tilde z_k' \tilde z_k \geq \varphi_{\min}(m + s_0)(G)^{2}$.  This is seen by noting that $\tilde z_k$ results in the composition of two projections onto a span of covariates of size bounded by $m + s_0$.  


This gives $$\frac{1}{n\tilde w_j' \tilde w_j}(\tilde \gamma_j ' \tilde \theta)^2 \geq \frac{1}{4} \frac{1}{n}\varphi_{\min}(m+s_0)(G)^{2} \tilde \theta_k^2.$$
Using the fact that $\tilde w_j' \tilde w_j \geq 1$ implies that

$$(\tilde \gamma_{j} ' \tilde \theta )^2/ \tilde \theta_k^2  \geq  \frac{1}{4} \varphi_{\min}(m + s_0)(G)^{2}.$$
Now suppose no true variables remain when $j$ is selected, then $\tilde w_j' \tilde w_j = \tilde u_j' \tilde u_j = 1$.  Therefore, 

$$-\Delta_{j} \ell(S_{\text{pre-}w_j}) =\frac{1}{n} \tilde \gamma_{j\tilde \eps}^2 \tilde \theta_{\tilde \eps}^2 \geq t$$
Note that $\tilde \theta$ is given by $\tilde \theta_{\tilde \eps} =\tilde \eps' y $ $=\eps ' \mathscr M_{s_0} y / (\eps' \mathscr M_{s_0} \eps)^{1/2} =  (\eps' \mathscr M_{s_0} \eps)^{1/2}. $
Therefore, $$ \tilde \gamma_{j\tilde \eps}^2 \geq t \frac{n}{\eps' \mathscr M_{s_0} \eps} .$$
This implies that $j \in A_2$.  Therefore, set $C_1 =  \frac{1}{2}\varphi_{\min}(m + s_0)(G)$.  

Next, construct $C_2$.  For each selected true covariate, $v_k$, set  $S_{\text{pre-}v_k}$ to be the set of (both true and false) covariates selected prior to $v_k$.   Note that $$ \tilde \theta_k^2  = - \Delta_k \ell(\{v_1,...,v_{k-1} \}) \geq - \Delta_k \ell(S_{\text{pre-}v_k}) $$  since $\{v_1,...,v_{k-1} \} \subseteq S_{\text{pre}-v_k}$.  In addition, if $v_k$ is selected before $v_l$ (or $v_l$ is not selected), then $$  - \Delta_k \ell(S_{\text{pre-}v_k}) \geq - \Delta_l \ell(S_{\text{pre-}v_k}) \geq \tilde z_l'\tilde z_l \tilde \theta_l ^2 \geq \varphi_{\min}(\hat s + s_0)(G)^{2} \tilde \theta_l^2.$$
Therefore, taking $$C_2 =  \varphi_{\min}(m + s_0)(G)$$ implies that $\tilde \theta_k /\ \tilde \theta_l \geq C_2$ for any $l > k$.

\

\subsection*{Step 6}   In this section, the number of elements of $A_2$ is bounded.  Recall that the criteria for $j\in A_2$ is that $|\tilde \gamma_{j \tilde \eps} |>  \frac{t^{1/2}n^{1/2}}{(2\eps'\mathscr M_{s_0} \eps )^{1/2}}  $.  Note also that $\tilde \gamma_{j\tilde \eps}$ is found by the coefficient in the expression



$$ \tilde \gamma_{j\tilde \eps} =    \tilde \eps' \tilde w_j = \eps'  \frac{1}{( \eps ' \mathscr M_{s_0} \eps)^{1/2}}   \mathscr M_{s_0} \tilde  w_j  $$




Next, let $H$ be the matrix $H = [v_1,...,v_{s_0}, w_1,...,w_m]$.  Note that $$\frac{1 }{( \eps ' \mathscr M_{s_0} \eps)^{1/2}}   \mathscr M_{s_0} \tilde  w_j   \in \text{span}(H)$$ 

Which implies that the above expression is unchanged when premultiplied by $H(H'H)^{-1}H'$.  Therefore, $$ \tilde \gamma_{j\tilde \eps} = \eps ' H (H'H)^{-1} H' \frac{1}{( \eps ' \mathscr M_{s_0} \eps)^{1/2} }  \mathscr M_{s_0} \tilde  w_j. $$
Let $\mu_j$ be the $+1$ for each $j \in A_2 $ such that $\tilde \gamma_{j\tilde \eps} > 0$ and $-1$ for each $j\in A_2$ such that $\tilde \gamma_{j\tilde \eps} < 0$. By the fact that $j \in A_2$, $\tilde \gamma_{j\tilde \eps}\mu_j > \frac{t^{1/2} n^{1/2} }{(2 \eps' \mathscr M_{s_0})^{1/2}}$ , summing over $j \in A_2$ gives $$\sum_{j \in A_{2}} \eps ' H (H'H)^{-1} H' \frac{1}{( \eps ' \mathscr M_{s_0} \eps)^{1/2}} \mathscr M_{s_0} \tilde  w_j  \mu_j > m_2 \frac{t^{1/2}n^{1/2}}{ (2\eps'\mathscr M_{s_0} \eps )^{1/2}} $$
This implies that 
$$\Big \| (H'H)^{-1} H' \frac{1 }{( \eps ' \mathscr M_{s_0} \eps)^{1/2}}  \sum_{j \in A_{2}} \mathscr M_{s_0} \tilde  w_j  \mu_j   \Big \|_1 \|\eps'H \|_\infty > m_2  \frac{t^{1/2}n^{1/2} }{ (2\eps'\mathscr M_{s_0} \eps )^{1/2}} $$
Which further implies that 
$$\sqrt{m+s_0} \Big \|  (H'H)^{-1} H' \frac{1}{( \eps ' \mathscr M_{s_0} \eps)^{1/2}}  \sum_{j \in A_{2}} \mathscr M_{s_0} \tilde  w_j  \mu_j    \Big \|_2 \|\eps'H \|_\infty > m_2 \frac{t^{1/2}n^{1/2}}{(2\eps'\mathscr M_{s_0} \eps )^{1/2}} $$
Next,  further upper bound the $\| \cdot \|_2$ term on the left side above by 


$$ \Big \|  (H'H)^{-1} H'\frac{1}{( \eps ' \mathscr M_{s_0} \eps)^{1/2}}   \sum_{j \in A_{2}} \mathscr M_{s_0} \tilde  w_j  \mu_j    \Big \|_2  $$ $$\leq  \frac{n^{-1/2} }{( \eps ' \mathscr M_{s_0} \eps)^{1/2}}   \varphi_{\min}(s_0+m)(G)^{-1/2}   \|  \mathscr M_{s_0} \sum_{j \in A_{2}} \tilde  w_j  \mu_j  \|_2$$  
next, by the fact that $\mathscr M_{s_0}$ is a projection (hence non-expansive)  and $\tilde w_j$ are mutually orthogonal, 
$$ \leq \frac{n^{-1/2}}{( \eps ' \mathscr M_{s_0} \eps)^{1/2}}   \varphi_{\min}(s_0+m)(G)^{-1/2} \sqrt{ \sum_{j \in A_{2}}  \| \tilde  w_j  \mu_j  \|_2^2} .$$
In Section 3, it was shown that $\max_j \| \tilde w_j\|_2^2 \leq \varphi_{\min}(s_0 + m)(G)^{-1}$.   Therefore, putting the above inequalities together, 
$$\frac{n^{-1/2} }{( \eps ' \mathscr M_{s_0} \eps)^{1/2}}  \sqrt{m+s_0} \varphi_{\min}(m+s_0)(G)^{-1} \sqrt{m_2}\| \eps'H\|_{\infty} > m_2  \frac{t^{1/2}n^{1/2}}{ (2\eps'\mathscr M_{s_0} \eps )^{1/2}}  .$$
This implies that 

$$m_2 < \frac{1}{n}\frac{2}{t} (\eps'\mathscr M_{s_0} \eps )(m+s_0)  \frac{ \| \eps'H\|_{\infty}^2}{ \eps ' \mathscr M_{s_0} \eps} \varphi_{\min}(m+s_0)(G)^{-2}$$
$$\leq 2(m +s_0)\frac{ \|  \En[ x_i \eps_i ]\|_{\infty}^2}{t } \varphi_{\min}(m+s_0)(G)^{-2}.$$
Under the assumed condition that $t^{1/2}  \geq 2 \| \En[x_i \eps_i ]\|_{\infty} \varphi_{\min}(m+s_0)(G)^{-1}$,  it follows that
$$m_2 \leq \frac{1}{2}(m+s_0).$$
By substituting $m = m_1 + m_2$ gives 
$m_2 \leq  m_1 + s_0.$

\subsection*{Step 7}This section concludes the proof of the second statement of the theorem by bringing together all of the facts proven in Steps 3-6. 
Combining 
$m_1 \leq   \varphi_{\min}(m+s_0)(G)^{-1}C_1^{-2} {C_3}^2 \(K_G^{\mathbb R} \)^2s_0 $
and
$m_2 \leq  m_1 + s_0$
gives 
$$m \leq  \[2  \varphi_{\min}(m+s_0)(G)^{-1} C_1^{-2} {C_3}^2 {K_G^{\mathbb R} }^2   
 + 1 \]s_0.$$  In addition, 
 \begin{align*} C_1 &= \frac{1}{2} \varphi_{\min}(m + s_0)(G), \\
C_2 &=  \varphi_{\min}(m + s_0)(G), \\  
C_3 &= (1 +  \varphi_{\min}(m + s_0)(G)^{-1/2} +  \varphi_{\min}(m + s_0)(G)^{-1}), 
\end{align*}
and $K_G^{\mathbb R} < 1.783$.  Therefore, 
\begin{align*} m & \leq \Big [ 1 + 8 \times 1.783^2  \times  \varphi_{\min}(m+s_0)(G)^{-3}\\
& \quad \times (1 +  \varphi_{\min}(m + s_0)(G)^{-1/2} +  \varphi_{\min}(m + s_0)(G)^{-1})^{2} \Big ] s_0.
\end{align*}

Since $C_3^2 \leq 9 \varphi_{\min}(s_0+m)(G)^{-2}$, the expression above can be simplified at the expense of a slightly less tight constant, so that
$$m \leq [ 1 + 72 \times 1.783^2  \times  \varphi_{\min}(m+s_0)(G)^{-5}   ]s_0.$$
Since this bound holds for each positive integer $m$ of wrong selections, this concludes the proof of Theorem 1.  

\end{proof}



\section{Proof of Theorems 2 and 3}
Theorem 2 follows by applying Theorem 1 in the following way.  If $\hat s$ grows faster than $s_0$, then there is $m< \hat s$ such that $s_0 < m < K_n$ and $m/s_0$ exceeds $\mathscr C_2(K_n) = O(1)$, giving a contradiction.  The first statement of the theorem follows from applying the bound on $\hat s$.  Theorem 3 follows by $\| \theta_0 - \thetahat \|_1 \leq \sqrt{\hat s + s_0} \| \theta_0 - \thetahat \|_2 \leq \sqrt{\hat s + s_0} \varphi_{\min}(\hat s + s_0)(G_x)^{-1} \En[(x_i'\theta_0 - x_i' \thetahat)^2]^{1/2}$.

\section{Conclusion}

This paper proved convergence rates for forward selection.  The rates in prediction error norm match typical rates encountered with other high-dimensional estimation techniques like Lasso.  The results are derived under minimal conditions which do not require $\beta$-min assumptions or irrepresentability.

\bibliographystyle{plain}
\bibliography{dkbib1}

\begin{thebibliography}{10}

\bibitem{BaiNg2008}
J.~Bai and S.~Ng.
\newblock Forecasting economic time series using targeted predictors.
\newblock {\em Journal of Econometrics}, 146:304--317, 2008.

\bibitem{BaiNg2009b}
J.~Bai and S.~Ng.
\newblock Boosting diffusion indices.
\newblock {\em Journal of Applied Econometrics}, 24, 2009.

\bibitem{BellChenChernHans:nonGauss}
A.~Belloni, D.~Chen, V.~Chernozhukov, and C.~Hansen.
\newblock Sparse models and methods for optimal instruments with an application
  to eminent domain.
\newblock {\em Econometrica}, 80:2369--2429, 2012.
\newblock Arxiv, 2010.

\bibitem{BC-PostLASSO}
A.~Belloni and V.~Chernozhukov.
\newblock Least squares after model selection in high-dimensional sparse
  models.
\newblock {\em Bernoulli}, 19(2):521--547, 2013.
\newblock ArXiv, 2009.

\bibitem{BCHK:Panel}
A.~Belloni, V.~Chernozhukov, C.~Hansen, and D.~Kozbur.
\newblock Inference in high dimensional panel models with an application to gun
  contorl.
\newblock {\em ArXiv:1411.6507}, 2014.

\bibitem{BCH-PLM}
Alexandre Belloni, Victor Chernozhukov, and Christian Hansen.
\newblock Inference on treatment effects after selection amongst
  high-dimensional controls with an application to abortion on crime.
\newblock {\em Review of Economic Studies}, 81(2):608--650, 2014.

\bibitem{BickelRitovTsybakov2009}
P.~J. Bickel, Y.~Ritov, and A.~B. Tsybakov.
\newblock Simultaneous analysis of {L}asso and {D}antzig selector.
\newblock {\em Annals of Statistics}, 37(4):1705--1732, 2009.

\bibitem{BuhlmannGeer2011}
P.~B\"{u}hlmann and S.~van~de Geer.
\newblock {\em Statistics for High-Dimensional Data: Methods, Theory and
  Applications}.
\newblock Springer, 2011.

\bibitem{BuhlmannBoosting2006}
Peter B\"uhlmann.
\newblock Boosting for high-dimensional linear models.
\newblock {\em Ann. Statist.}, 34(2):559--583, 2006.

\bibitem{BuneaTsybakovWegkamp2007b}
F.~Bunea, A.~Tsybakov, and M.~H. Wegkamp.
\newblock Sparsity oracle inequalities for the lasso.
\newblock {\em Electronic Journal of Statistics}, 1:169--�194, 2007.

\bibitem{BuneaTsybakovWegkamp2006}
F.~Bunea, A.~B. Tsybakov, , and M.~H. Wegkamp.
\newblock Aggregation and sparsity via $\ell_1$ penalized least squares.
\newblock In {\em Proceedings of 19th Annual Conference on Learning Theory
  (COLT 2006) (G. Lugosi and H. U. Simon, eds.)}, pages 379--�391, 2006.

\bibitem{BuneaTsybakovWegkamp2007}
F.~Bunea, A.~B. Tsybakov, and M.~H. Wegkamp.
\newblock Aggregation for {G}aussian regression.
\newblock {\em The Annals of Statistics}, 35(4):1674--1697, 2007.

\bibitem{CandesTao2007}
E.~Cand\`{e}s and T.~Tao.
\newblock The {D}antzig selector: statistical estimation when $p$ is much
  larger than $n$.
\newblock {\em Ann. Statist.}, 35(6):2313--2351, 2007.

\bibitem{Submodular:Spectral}
Abhimanyu Das and David Kempe.
\newblock Submodular meets spectral: Greedy algorithms for subset selection,
  sparse approximation and dictionary selection.
\newblock In Lise Getoor and Tobias Scheffer, editors, {\em Proceedings of the
  28th International Conference on Machine Learning (ICML-11)}, pages
  1057--1064, New York, NY, USA, 2011. ACM.

\bibitem{EfronHastieJohnstoneTibshirani2004}
B.~Efron, T.~Hastie, I.~Johnstone, and R.~Tibshirani.
\newblock Least angle regression.
\newblock {\em Annals of statistics}, 32(2):407--451, 2004.

\bibitem{FF:1993}
Ildiko~E. Frank and Jerome~H. Friedman.
\newblock A statistical view of some chemometrics regression tools.
\newblock {\em Technometrics}, 35(2):109--135, 1993.

\bibitem{Freund+Schapire:1996}
Yoav Freund and Robert~E. Schapire.
\newblock Experiments with a new boosting algorithm.
\newblock In {\em Proceedings of the Thirteenth International Conference on
  Machine Learning (ICML 1996)}, 1996.

\bibitem{Vershynin:Le:GraphConcentration}
O.~{Gu{\'e}don} and R.~{Vershynin}.
\newblock {Community detection in sparse networks via Grothendieck's
  inequality}.
\newblock {\em ArXiv e-prints}, November 2014.

\bibitem{elements:book}
Trevor Hastie, Robert Tibshirani, and Jerome Friedman.
\newblock {\em Elements of Statistical Learning: Data Mining, Inference, and
  Prediction}.
\newblock Springer, New York, NY, 2009.

\bibitem{horowitz:lasso}
Jian Huang, Joel~L. Horowitz, and Fengrong Wei.
\newblock Variable selection in nonparametric additive models.
\newblock {\em Ann. Statist.}, 38(4):2282--2313, 2010.

\bibitem{knight:shrinkage}
Keith Knight.
\newblock Shrinkage estimation for nearly singular designs.
\newblock {\em Econometric Theory}, 24:323--337, 2008.

\bibitem{Koltchinskii2009}
V.~Koltchinskii.
\newblock Sparsity in penalized empirical risk minimization.
\newblock {\em Ann. Inst. H. Poincar� Probab. Statist.}, 45(1):7--57, 2009.

\bibitem{TBFMS1}
D.~{Kozbur}.
\newblock {Testing-Based Forward Model Selection}.
\newblock {\em ArXiv e-prints}, December 2015.

\bibitem{Lounici2008}
K.~Lounici.
\newblock Sup-norm convergence rate and sign concentration property of lasso
  and dantzig estimators.
\newblock {\em Electron. J. Statist.}, 2:90--102, 2008.

\bibitem{LouniciPontilTsybakovvandeGeer2009}
K.~Lounici, M.~Pontil, A.~B. Tsybakov, and S.~van~de Geer.
\newblock Taking advantage of sparsity in multi-task learning.
\newblock {\em arXiv:0903.1468v1 [stat.ML]}, 2010.

\bibitem{MRY2007}
N.~Meinshausen, G.~Rocha, and B.~Yu.
\newblock Discussion: A tale of three cousins: Lasso, l2boosting and dantzig.
\newblock {\em Ann. Statist.}, 35(6):2373--2384, 2007.

\bibitem{MY2007}
N.~Meinshausen and B.~Yu.
\newblock Lasso-type recovery of sparse representations for high-dimensional
  data.
\newblock {\em Annals of Statistics}, 37(1):2246--2270, 2009.

\bibitem{RosenbaumTsybakov2008}
M.~Rosenbaum and A.~B. Tsybakov.
\newblock Sparse recovery under matrix uncertainty.
\newblock {\em arXiv:0812.2818v1 [math.ST]}, 2008.

\bibitem{RudelsonZhou2011}
M.~Rudelson and S.~Zhou.
\newblock Reconstruction from anisotropic random measurements.
\newblock {\em ArXiv:1106.1151}, 2011.

\bibitem{RudelsonVershynin2008}
Mark Rudelson and Roman Vershynin.
\newblock On sparse reconstruction from fourier and gaussian measurements.
\newblock {\em Communications on Pure and Applied Mathematics}, 61:1025�1045,
  2008.

\bibitem{tao:blog:transitivity}
T.~Tao.
\newblock When is correlation transitive?
  [https://terrytao.wordpress.com/2014/06/05/when-is-correlation-transitive/],
  June 2014.

\bibitem{T1996}
R.~Tibshirani.
\newblock Regression shrinkage and selection via the lasso.
\newblock {\em J. Roy. Statist. Soc. Ser. B}, 58:267--288, 1996.

\bibitem{Tropp:Greed}
Joel~A. Tropp.
\newblock Greed is good: algorithmic results for sparse approximation.
\newblock {\em Information Theory, IEEE Transactions on}, 50(10):2231--2242,
  Oct 2004.

\bibitem{vdGeer}
S.~A. van~de Geer.
\newblock High-dimensional generalized linear models and the lasso.
\newblock {\em Annals of Statistics}, 36(2):614--645, 2008.

\bibitem{vdGBRD:AsymptoticConfidenceSets}
Sara van~de Geer, Peter Bühlmann, Ya’acov Ritov, and Ruben Dezeure.
\newblock On asymptotically optimal confidence regions and tests for
  high-dimensional models.
\newblock {\em Ann. Statist.}, 42(3):1166--1202, 06 2014.

\bibitem{Wainright2006}
M.~Wainwright.
\newblock Sharp thresholds for noisy and high-dimensional recovery of sparsity
  using $\ell_1$-constrained quadratic programming (lasso).
\newblock {\em IEEE Transactions on Information Theory}, 55:2183--2202, May
  2009.

\bibitem{Wang:UltraHighForwardReg}
Hansheng Wang.
\newblock Forward regression for ultra-high dimensional variable screening.
\newblock {\em Journal of the American Statistical Association},
  104:488:1512--1524, 2009.

\bibitem{Yao2007:EarlyStopping:GradDesc}
Yuan Yao, Lorenzo Rosasco, and Andrea Caponnetto.
\newblock On early stopping in gradient descent learning.
\newblock {\em Constructive Approximation}, 26(2):289--315, 2007.

\bibitem{ZhangHuang2006}
C.-H. Zhang and J.~Huang.
\newblock The sparsity and bias of the lasso selection in high-dimensional
  linear regression.
\newblock {\em Ann. Statist.}, 36(4):1567--1594, 2008.

\bibitem{Zhang:GreedyLeastSquares}
Tong Zhang.
\newblock On the consistency of feature selection using greedy least squares.
\newblock {\em Journal of Machine Learning}, 10:555--568, 2009.

\bibitem{TongZhang:ForwardBackward}
Tong Zhang.
\newblock Adaptive forward-backward greedy algorithm for learning sparse
  representations.
\newblock {\em IEEE Trans. Inf. Theor.}, 57(7):4689--4708, July 2011.

\bibitem{zhang2005}
Tong Zhang and Bin Yu.
\newblock Boosting with early stopping: Convergence and consistency.
\newblock {\em Ann. Statist.}, 33(4):1538--1579, 08 2005.

\end{thebibliography}

\end{document}